\documentclass{article}
\usepackage{spconf}
\usepackage{times}
\usepackage{latexsym}
\usepackage{amsmath}
\usepackage{multirow}
\usepackage{url}
\usepackage[normalem]{ulem}
\usepackage{color}
\usepackage[usenames,dvipsnames]{xcolor}
\usepackage{enumitem}
\usepackage{graphicx}
\usepackage{bibspacing}
\name{Lu Wang$^{\star}$, Larry Heck$^{\dagger}$, Dilek Hakkani-T\"ur$^{\dagger}$}
\address{$^{\star}$Cornell University, Department of Computer Science, Ithaca, NY, USA\\
$^{\dagger}$Microsoft Research, Mountain View, CA, USA}

\begin{document}
\ninept

\title{Leveraging Semantic Web Search and Browse Sessions for Multi-Turn Spoken Dialog Systems}

\maketitle

\begin{abstract}
Training statistical dialog models in spoken dialog systems (SDS) requires large amounts of annotated data. The lack of scalable methods for data mining and annotation poses a significant hurdle for state-of-the-art statistical dialog managers. This paper presents an approach that directly leverage billions of web search and browse sessions to overcome this hurdle. The key insight is that task completion through web search and browse sessions is (a) predictable and (b) generalizes to spoken dialog task completion. The new method
automatically mines behavioral search and browse patterns from web logs and translates them into spoken dialog models. We experiment with naturally occurring spoken dialogs and large scale web logs. Our session-based models outperform the state-of-the-art method for entity extraction task in SDS. We also achieve better performance for both entity and relation extraction on web search queries when compared with nontrivial baselines.
\end{abstract}

\begin{keywords}
spoken dialog systems, statistical dialog management, multi-turn contextual model, dialog session, web search session, web browsing, entity extraction, relation extraction
\end{keywords}

\section{Introduction}
Statistical machine learning approaches for spoken dialog systems (SDS) have become popular in the last decade. However, an important problem that limits their practical application is the significant amount of annotated data that is required for training models in SDS~\cite{Suendermann_Pieraccini_2011}. This is particularly true for the dialog manager (DM) component, where the number of possible dialog state sequences is extremely high and difficult to enumerate. Hence scaling SDS approaches to handle even moderately complex dialog models is a key research problem.

A simple approach to solve this problem relies on a flat initialization; as the system is used, training data is obtained from real users and annotated to train better dialog models. However, such a bootstrapped statistical model is not desirable for real-world SDS as the initial user experience is poor and limited. In addition, the subsequent learning is biased towards simplified interactions, since these are the only dialogs that yield success for the user.

A more common solution is to exploit a small set of real user interactions to perform dialog simulations. With this approach, one can obtain a large set of artificial dialogs~\cite{Scheffler:2002:ALD:1289189.1289246,Williams:2007:POM:1221595.1221967,Lemon07machinelearning,GriolRS09}. Though simple to implement, the generated dialogs are typically not natural, and do not provide the opportunity for the system to learn the best ways to complete tasks from users. It may result in an SDS that leads a user through inefficient and occasionally non-sensical paths, causing user frustration and ultimately poor adoption rates~\cite{DBLP:journals/ker/FramptonL09}.

\begin{figure}[htbp]
\scriptsize
\begin{tabular}{l}
--- $\bf S_{1}: $ \underline{\textbf{Web Search and Browse Session}}\\
$T_{1}$ (Q): [the great gatsby]$_{\textbf{\color{blue}film}}$ 2013\\
$T_{2}$ (C): [\url{www.imdb.com/title/tt1343092/}]$_{\textbf{\color{blue}film}}$\\
$T_{3}$ (Q): movies by [leonardo dicaprio]$_{\textbf{\color{blue}actor}}$\\
$T_{4}$ (C): [\url{www.imdb.com/name/nm0000138/}]$_{\textbf{\color{black}actor}}$\\
$T_{5}$ (Q): [titanic]$_{\textbf{\color{blue}film}}$\\
$T_{6}$ (C): [\url{www.imdb.com/title/tt0120338/}]$_{\textbf{\color{black}film}}$\\
$T_{7}$ (C): [\url{http://www.imdb.com/character/ch0002338/}]$_{\textbf{\color{black}character}}$\\
$T_{8}$ (Q): does [leonardo dicaprio]$_{\textbf{\color{black}actor}}$ have an award\\
\end{tabular}
\\~\\
\begin{tabular}{l}
--- $\bf S_{2}: $ \underline{\textbf{Spoken Dialog System Session}}\\
$U_{1}$: go to bing dot com\\
$U_{2}$: search for [the great gatsby]$_{\textbf{\color{blue}film}}$\\
$U_{3}$: click that ([www.imdb.com/title/tt1343092/]$_{\textbf{\color{blue}film}}$)\\
$U_{4}$: show me movies by [leonardo dicaprio]$_{\textbf{\color{blue}actor}}$\\
$U_{5}$: scroll down\\
$U_{6}$: show me information on [titanic]$_{\textbf{\color{blue}film}}$\\
$U_{7}$: go back\\
$U_{8}$: show me information on [inception]$_{\textbf{\color{black}film}}$\\
\end{tabular}
\vspace{-3mm}
\caption{\small Example fragments of a web search and browse session (S$_{1}$) and a spoken dialog system session (S$_{2}$). $T_{i}$/$U_{i}$ represents the $i$th turn/utterance in web search/dialog. For the web session, $Q$ or $C$ in the parentheses indicates if the current turn is a query or click. For the dialog session, we also include the URL the user refers to (e.g. $U_{3}$). Each entity in the query or utterance is shown within brackets along with its entity type in subscript.}
\label{session_fig}
\end{figure}

\noindent \textbf{Leveraging web search log to SDS.} 
Nonetheless, large volumes of task-oriented and potentially relevant training data exist in web search engines. Users complete tasks everyday on the web, using a combination of search and browsing. The volume of the search data easily exceeds 100M queries per day over hundreds of millions of users. The breadth of tasks is also immense, ranging from simple (weather, finding directions, local events) to more complex (shopping, planning a trip, planning a night out). 

In many ways, search and browsing have elements of automated conversational interactions. 
Figure~\ref{session_fig} displays one fragment of web search session, and one fragment of an spoken dialog (conversational) session. We can see that the two sessions share similar browsing patterns (in \textbf{\color{blue}blue}) in terms of the entity types visited. Web search conversations are also interactive because the system responds to what has previously been communicated. The conversations are spontaneous as the user is not constrained by domain. 

The advantage of using web data is the massive volume of interactions, with both depth in the number and complexity of user interactions and breadth in the variety of user goals. Some of the user's actions (e.g., clicks) further provide enough constraints so that the labels (domains, intents, slots) can be {\it inferred}, enabling weakly supervised training. The observation thus inspires us to leverage the large volumes of logged user interactions that exist in centrally hosted web search engines and browsers to the SDS systems. 

In this work, we take a first step to effectively exploit this source of training data for an SDS problem -- entity extraction from the spoken dialogs. 

\noindent \textbf{Session-based model for semantic parsing in SDS.} 
Existing work for leveraging web search behavior to SDS mainly exploits offline information in the query-click logs. For instance, the web query-click graph has been employed for domain detection~\cite{HakkaniEtAl2011_asru}, slot filling \cite{conf/interspeech/TurHHC11}, and intent detection~\cite{HeckEtAl2012_slt}. Recently, the idea of combining the power of internet browser and multi-modal input (e.g. speech, gesture, etc) is discussed in~\cite{Heck2012_slt} and implemented by~\cite{HeckEtAl2013_slam}. 
Different from all of the work above, our session models are \textit{directly trained} on the web search sessions with contextual information from the search logs. We then test the models on entity extraction problem for SDS with the acquired behaviorial patterns from web and multi-turn conversational context. Authors in~\cite{conf/icassp/BhargavaHT13} also try to leverage contextual information for the same task, however, they do not exploit web data. 


\noindent \textbf{Distant supervision for query interpretation.} 
In addition, we investigate a domain-independent framework that utilizes a structured semantic knowledge base of large scale to provide distant supervision. Modern systems of entity and relation extraction from web search queries are mostly based on supervised learning with small hand-labeled corpora, and are thus limited by the availability of training data~\cite{Pound:2012:IKQ:2396761.2396803}. 
Weakly-supervised approaches are also studied~\cite{Riloff:1999:LDI:315149.315364,Pasca:2007:WDN:1321440.1321536}, where seed instances are utilized for boostrap learning. Noisy extractions, with semantic shift, are hard to avoid. 

With the emergence of large scale knowledge resources (e.g. Freebase, DBPedia), \textit{distant supervision} has been proposed for relation extraction on formal text, such as Wikipedia articles~\cite{Mintz:2009:DSR:1690219.1690287}, and news articles~\cite{Yao:2010:CCR:1870658.1870757}. 
It is also investigated in~\cite{TurEtAl2012_interspeech,HakkaniEtAl2013_icassp} for utilizing web search results to improve the relation extraction performance in SDS. In this work, we employ distant supervision to generate training samples based on web search queries instead of web pages returned by search engine. We also test \textit{entity extraction} and \textit{relation extraction} tasks on both spoken utterances and web search queries.

\noindent \textbf{Contributions.} 
To the best of our knowledge, we are the first to exploit web search sessions and their contextual information and adapt it to train statistical dialog managers. Our experiments conducted in the film domain show that our session-based models trained on web search log are capable to leverage multi-turn context to produce significantly better entity extraction performance than the state-of-the-art approach~\cite{HeckEtAl2013_interspeech} in SDS. 

Furthermore, to address the issue of lack of training data, we propose a distantly supervised approach to semantic parsing, by employing massive unlabeled web search logs combined with large dictionaries of entities and relations gathered from knowledge bases. We train linear-chain Conditional Random Fields~\cite{Lafferty:2001:CRF:645530.655813} and its variant learned on data with missing labels~\cite{citeulike:1773161} to model web browsing behavior on session level. Our models achieve better results than nontrivial baselines on entity extraction and relation extraction for both spoken dialogs and web search query. Notice that with knowledge base as supervision, entities extracted by our models are beyond named entity slots used in conventional SDS. We are also able to find unnamed categories, such as film subject and descriptions.



\section{The Query Interpretation Model}
Investigations from \cite{Pound:2012:IKQ:2396761.2396803} discover that three templates (or types) of queries dominate entity-based web search (i.e. more than 65\% of the sampled queries fall in these three types). The three templates can be formalized as: \texttt{Entity} (\textbf{E}), which contains one entity; \texttt{Type $\sqcap$ Relation(Entity)} (\textbf{T-R-E}), which describes one entity and another entity type with a certain relation; \texttt{Entity$_{0}$ $\sqcap$ Relation(Entity$_{1}$)} (\textbf{E-R-E}), which consists of a pair of related entities. In spoken dialog corpora studied in this work, we also observe more than 50\% of utterances can be matched to those templates, where 54\% of them belong to template \textbf{E}. 
For instance, in Figure~\ref{session_fig}, query ``search for the great gatsby" ($U_{2}$) falls into template \textbf{E} as it has a single film entity. Query ``show me movies by leonardo dicaprio" ($U_{4}$) matches template \textbf{T-R-E} where ``movies" indicates the film type and ``leonardo dicaprio" is an actor entity. A sample query for \textbf{E-R-E} could be ``Titanic by James Cameron".

For the remaining part of the section, we describe our full query interpretation pipeline, consisting of \textit{Entity Mention Identification}, \textit{Entity Type Determination}, and \textit{Relation Extraction}. 


\subsection{Entity Mention Identification}
\label{entMention}


We first employ a linear-chain Conditional Random Fields (CRF)~\cite{Lafferty:2001:CRF:645530.655813} to identify all the entity mentions in each query. 
We represent each query as a sequence of tokens, $\textbf{x}=x_{0}x_{1}\ldots x_{n}$, and generate a sequence of labels, $\textbf{y}=y_{0}y_{1}\ldots y_{n}$, that encode which tokens are part of entity mentions by using a BIO label format: \{\texttt{B-ENT} (beginning of an entity), \texttt{I-ENT} (inside of an entity), \texttt{O} (outside of any entity)\}. 
As defined in~\cite{Lafferty:2001:CRF:645530.655813}, the conditional probability is given by:\\

\vspace{-2mm}
$P(\textbf{y}|\textbf{x})= \frac{1}{Z(\textbf{x})} \exp (\sum_{t=1}^{T}\sum_{k=1}^{K}\lambda_{k}f_{k}(y_{t-1},y_{t},\textbf{x}, t))$\\

\vspace{-2mm}
\noindent where $T$ is the length of the sequence, $K$ is the number of feature functions $f_{k}$, and $\lambda_{k}$ are the corresponding weights. 

\noindent \textbf{Feature Design.}
We utilize lexical features such as \texttt{word in lowercase}, \texttt{is capitalized}, \texttt{has any number}, and \texttt{has any non-alphabetic character}. For syntactic features, \texttt{POS tag}, \texttt{NER tag}, and \texttt{dependency relation} output by Stanford parser~\cite{Klein:2003:AUP:1075096.1075150} are used. We also consider context features within the session. Figure~\ref{session_fig} shows that ``leonardo dicaprio" is mentioned twice in $T_{3}$ and $T_{8}$. Intuitively, if an ngram appears repeatedly in the queries, it is likely to be an entity mention. Thus, we use \texttt{repeat time} to indicate if an ngram is repeated more than twice in recent 5 issued queries within the session.

In addition, we collect \textit{gazetteers} from Freebase (\url{http://www.freebase.com/}), and use a feature to indicate if the token is part of a term in the gazetteer. Gazetteers, which are lists of entity names or relation instances, are important features for SDS to understand users' intent or represent dialog states. 
We then match each entity name in the gazetteer to the query. If there is any overlapping between two mapped names, e.g. ``leonardo dicaprio" v.s. ``dicaprio", the longer one is retained.

Suppose $t$ represents the current token position and $feature_{k}[t]$ is the value of $feature_{k}$, we extract unigram, bigram and trigram features over local context, e.g. $feature_{k}[t-2]$, $feature_{k}[t-2]:feature_{k}[t-1]$, $feature_{k}[t-2]:feature_{k}[t-1]:feature_{k}[t]$.

\vspace{-2mm}
\subsection{Entity Type Determination}
\label{entityTypeDeter}

\begin{figure}[ht]
\vspace{-5mm}
\includegraphics[width=3.6in,height=0.7in]{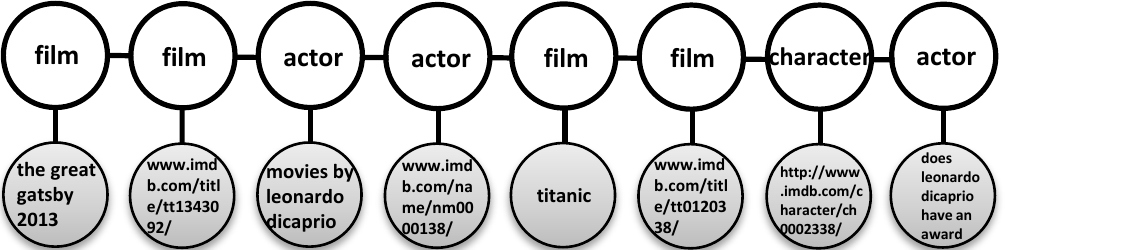}
\vspace{-8mm}
\caption{\small Example of linear-chain CRF for entity type determination.}
\label{entType_fig}
\end{figure}

Entity type is determined after identifying entity mentions, e.g. whether the entity mention is an ``actor" or a ``film". Instead of making prediction on each entity independent of the browsing history, we propose to use sequential labeling technique to predict entity types within a session simultaneously.

Suppose session $S$ consists of turns $\{s_{0}, s_{1}, \cdots, s_{m}\}$, where each turn $s_{i}$ can be either a query or a web page click. Each entity mentioned in $S$ represents a single state in our session-based entity type determination model. A click is mapped to an entity, if this URL is a source URL for that entity in Freebase. Figure~\ref{entType_fig} depicts a mapping from the web session in Figure~\ref{session_fig} to our sequential model.

\noindent \textbf{CRF Training with Missing Labels.} 
Linear-chain CRF as defined in Section~\ref{entMention} is utilized as the building block to predict entity types. 
Multiple entities in one query are considered as separate states.
During training, it is possible that some turns do not contain any entity, which makes the labels partially observed for the session. We propose two ways to address this problem: 1) simply removing those turns in training; 2) training a CRF with missing labels.

Given a session with partial labels $\textbf{y}=<y_{t_{0}}, y_{t_{1}}, \cdots, y_{t_{m}}>$, where $t_{i}$ is the index of an entity with observed label, and a hidden label sequence $\textbf{h}= <h_{l_{0}}, h_{l_{1}}, \cdots, h_{l_{m'}}>$, where $l_{j}$ is the index of an entity with latent (or unobserved) label, the conditional probability is,\\

\vspace{-2mm}
\noindent {\footnotesize $P(\textbf{y}|\textbf{x})= \sum_{\textbf{h}} p(\textbf{y}, \textbf{h}\mid x) =\frac{1}{Z(\textbf{x})}\sum_{\textbf{h}} \exp (\sum_{t=1}^{T}\sum_{k=1}^{K}\lambda_{k}f_{k}(y_{t-1},y_{t},\textbf{x}, t))$}\\

\vspace{-2mm}

We adopt the direct gradient ascent procedure methods from~\cite{citeulike:1773161} to maximize $P(\textbf{y}|\textbf{x})$. We omit details about parameter learning due to length limit, and refer the readers to~\cite{citeulike:1773161} for further information.

\noindent \textbf{Feature Design.} 
We use basic features such as \texttt{NER tag}, \texttt{is capitalized}, \texttt{number of tokens}, \texttt{URL domain}, and \texttt{is in gazetteer}. We also encode the context (i.e. ngrams and their POS tag sequence, $n\leq3$) of the entity mention as features. Likewise, unigram, bigram, trigram features are used as Section~\ref{entMention}.

\subsection{Relation Extraction}
After extracting entities from previous steps, we aim to identify queries satisfying templates \textbf{T-R-E} and \textbf{E-R-E} by using relation extraction techniques. For query with more than one entity, we predict if there exists a relation between pairwise entities (\textbf{E-R-E}). For entities not classified as \textbf{E-R-E}, we then predict if the query fits \textbf{T-R-E}. Specifically, for both templates, we train two multiclass logistic regression classifiers for relations based on one v.s. all strategy.

\noindent \textbf{Feature Design.} 
Our features are based on standard lexical and shallow syntactic features (i.e. POS tags, dependency relations) from the literature~\cite{Mintz:2009:DSR:1690219.1690287}. We also consider the words and their POS tags between the entity pairs or entity-type pairs. A gazetteer feature is constructed to indicate if the sample is in Freebase. We also take the predicted entity types from previous module as features.

\section{Web Search Query Interpretation}
\label{webSearch}

\noindent \textbf{Web Search Session Log (Search Log).}
We collected 12 weeks of web search sessions from March 18 to June 9, 2013. The sessions consisting of query-click logs based on IE browsing history. This dataset will be called ``\textbf{Search log}" henchforth.
We choose a representative domain -- \texttt{film}, to test our semantic parsing models. We further extract sessions that users express interests in film, i.e. sessions with page views in the domain of ``\url{www.imdb.com}", ``\url{www.fandango.com}", or ``\url{www.rottentomatoes.com}". This results in 297,352 sessions in total, from which 80\% are randomly selected for training, with the remaining as the test set. The same tasks for other domains will be studied in the future work.

\noindent \textbf{Training via Distant Supervision.} 
To construct training samples for entity extraction, we first collect a gazetteer consisting of entities in the form of names from Freebase along with their entity types. In the case that one entity has multiple types, the type having most instances in Freebase is chosen.
To maintain high precision for the training set, we match entities with names longer than two words in the gazetteer to each query. Table~\ref{match_stat_tab} shows nearly half of the queries and URLs get matched with at least one entity.

A gazetteer is also constructed for relation extraction. If a query contains two entities that compose a relation instance in Freebase, we assume this query expresses that relation. Those queries make up the training set for template \textbf{E-R-E}. For template \textbf{T-R-E}, a query is a positive sample when it contains entity $e$ (e.g. ``leonardo dicaprio") and description for entity type $t$ (e.g. ``movies"), and there exists entity $e^{\prime}$ (e.g. ``Titanic") of type $t$ where $e$ and $e^{\prime}$ carry a relation.

Table~\ref{expsetup_stat_tab} lists the number of entity and relation instances matched in our training and test sets.

\vspace{-3mm}
\begin{table}[ht]
\centering
\footnotesize
\begin{tabular}{|c|c|c|c|}
\hline
 & \textbf{Total} & \textbf{Matched} & \textbf{Matched \%}\\ \hline
Query & 1,773,930 &  866,131 & 48.83\% \\ \hline
URL & 18,166,912 &  8,413,971 & 46.31\% \\ \hline
\end{tabular}
\vspace{-3mm}
\caption{\small Matching statistics for Search log.}
\label{match_stat_tab}
\end{table}

\vspace{-7mm}
\begin{table}[ht]
\centering
\footnotesize
\begin{tabular}{|l|c|c|c|c|}
\hline
& \textbf{\# Ent Type} & \textbf{\# Ent} & \textbf{\# Rel Type} & \textbf{\# Rel} \\ \hline
Training & 44 & 215,053& 33& 20,197\\ \hline
Test & 44 & 104,657 & 27 & 5,255 \\ \hline
Test Only & 0& 23,557& 0 & 1,104\\ \hline
\end{tabular}
\vspace{-3mm}
\caption{\small Statistics for \textbf{Training} and \textbf{Test} dataset. \textbf{Test Only} denotes the instance (or type) in the test set is not observed in training.}
\label{expsetup_stat_tab}
\end{table}
\vspace{-5mm}

\subsection{Pilot Study: Entity-based Search Session Modeling}

We first carry out experiments to validate our session assumption, i.e. current search behavior depends on the searching history. Formally, in our entity-based session modeling, a search session $S$ is represented by a sequence of entities $e_{0}, e_{1}, \cdots, e_{l}$, as described in Section~\ref{entityTypeDeter}. We compute the likelihood of the sessions based on Markov models with different history, where higher order model considers longer history. 
The likelihood of session $S$ is expressed as the product of the entities that compose $S$, with each entity probability conditional on the previous $n-1$ entities (i.e. order-$n$): {\footnotesize $P(S)=\prod_{i=1}^{l}P(e_{i}|e^{i-1}_{i-n+1})$}, where $e^{j}_{i}$ denotes the sequence $e_{i},...,e_{j}$.


%
%

Maximum likelihood estimation can be used to get {\footnotesize $P_{ML}(e_{i}|e^{i-1}_{i-n+1})$} as {\footnotesize $\frac{c(e^{i}_{i-n+1})}{c(e^{i-1}_{i-n+1})}$}, where $c(\alpha)$ indicates the frequency of entity sequence $\alpha$ occurs. We also experiment with two smoothing techniques: \textit{additive smoothing} and \textit{linear interpolation}~\cite{Chen:1996:ESS:981863.981904}. Due to space limit, we refer the readers to~\cite{Chen:1996:ESS:981863.981904} for details. 
%
%
%
%
%
To handle unknown entities in test set, we replace the first appearance of each entity with label ``$<$Unk$>$" during training. Unknown entities are also substituted with ``$<$Unk$>$" during test.

Given test set $T=\{S_{i}\}$, where $S_{i}$ represents a session, we take the performance of each model as its cross-entropy~\cite{Chen:1996:ESS:981863.981904} on $T$, i.e. {\footnotesize $\frac{1}{N} \sum_{i=1}^{\mid T \mid} -log P(S_{i})$}, 
with $N$ as the total number of entities.

\begin{figure}[ht]
\hspace{10mm}
\includegraphics[width=2.6in,height=1.5in]{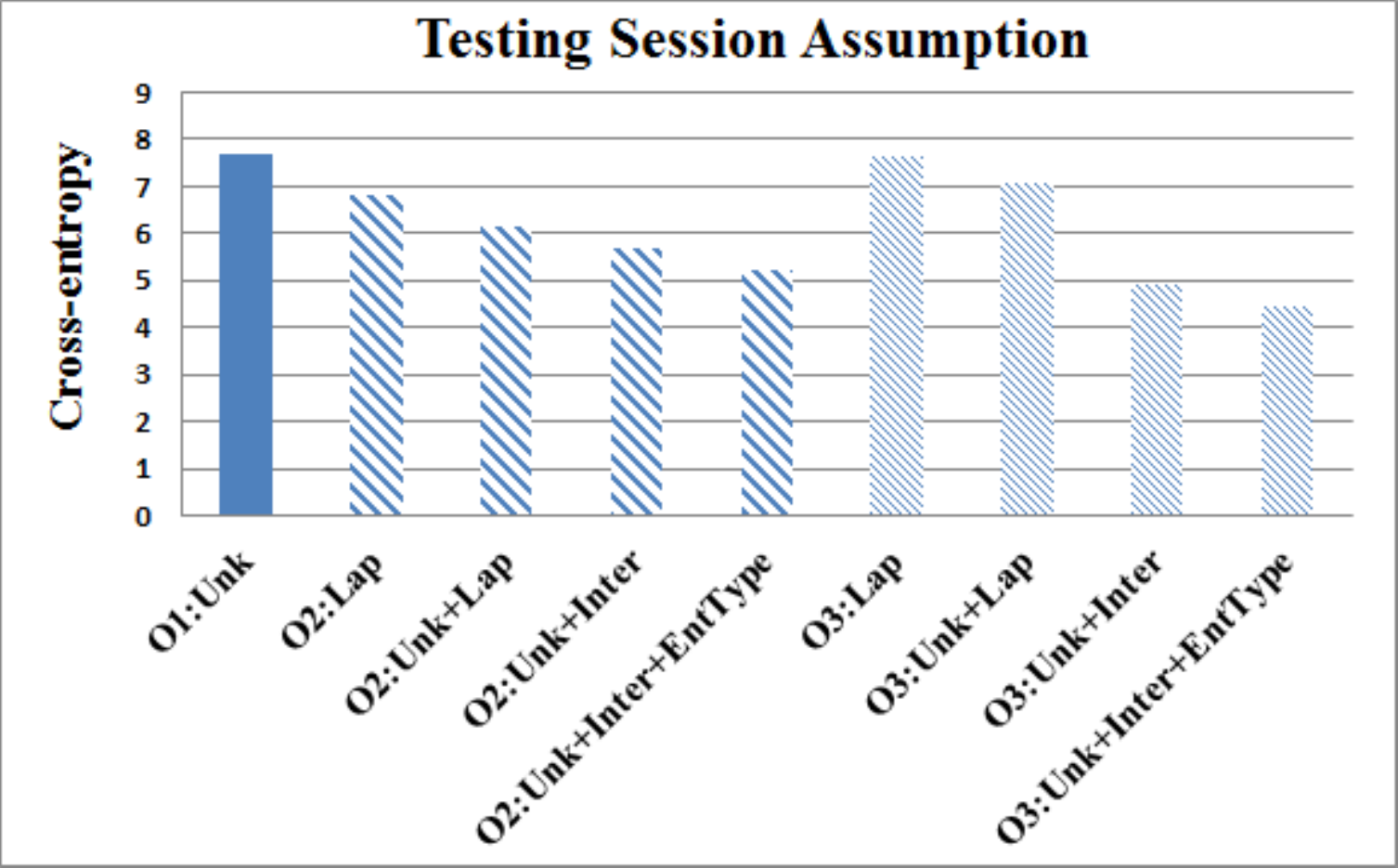}
\vspace{-4mm}
\caption{\small We compare models with different history -- order-1 (\textbf{O1}), order-2 (\textbf{O2}), and order-3 (\textbf{O3}). Models with unknown entity handling are labeled with ``\textbf{Unk}". Two smoothing techniques are studied: additive smoothing (a.k.a \textbf{Lap}lace smoothing) and linear interpolation (\textbf{Inter}). For interpolation, we also try models with entity type (``\textbf{EntType}") as part of the components, i.e. adding $P_{ML}(e_{i}|type(e)^{i-1}_{i-n+1})$, which produce the best performance (i.e. lowest cross-entropy) combined with longer history.}
\label{ngram_fig}
\end{figure}

We train each model on sessions collected from May 13, 2013 to May 19, 2013, and test on sessions of June 17, 2013 to June 23, 2013. We reserve 10\% of the training data as development set to tune the parameters. Results in Figure~\ref{ngram_fig} show that Markov models considering longer search history produce better performance (i.e. with lower cross-entropy), which confirms our session assumption.

\begin{table}[ht]
\centering
\footnotesize
\begin{tabular}{|l|c|c|}
\hline
& \textbf{Recall on E-R-E} & \textbf{Recall on T-R-E}\\\hline
Baseline (Gazetteer Only) & 79.21 & 74.72\\\hline
Our model & \textbf{80.81} & \textbf{78.22}\\\hline
\end{tabular}
\vspace{-3mm}
\caption{\small Results for relation extraction on queries from Search log.}
\label{web_relation_tab}
\vspace{-4mm}
\end{table}

\begin{table*}[htbp]
\centering
\footnotesize
\begin{tabular}{|l|c|c|c|c|c|c|c|c|c|c|}
\hline
& \multicolumn{5}{|c|}{\textbf{CB Data}} & \multicolumn{5}{|c|}{\textbf{Netflix Data}}\\ \hline
& Film & Director & Actor & Character & Genre & Film & Director & Actor & Character & Genre\\ \hline
(\# utterances) & 195& 16& 55& 3& 58& 2,013 & 196& 713& 85&968\\ \hline
Baseline 1 (Gazetteer Only) & 56.61 & 89.66 & 93.20 & 50.00 & 18.75 & 58.55 & 91.83 & 92.82 & 68.22 & 18.01 \\ \hline
Baseline 2 (No Session) & 71.71& 81.25 & 90.57& 50.00 & 18.75 & 67.98 & 75.29 & 89.23 & 70.23 & 18.01 \\ \hline
Heck et al.~\cite{HeckEtAl2013_interspeech} & -& -& -& -& -& 71.72 & 84.62& 58.61& -& 29.55\\ \hline
Session model (CRF) & 71.71 & 89.66 & 95.24$\dagger$ & 50.00 & 18.75 & 67.98 & 91.83 & 95.12$\dagger$ & 70.23 & 29.29$\dagger$ \\ \hline
Session model (CRF with missing label) &  \textbf{75.80} & 89.66 & 96.23$\dagger$ & 80.00 & \textbf{21.54} & \textbf{69.39} & 92.01 & 95.12$\dagger$ & 70.23 & \textbf{33.70}\\ \hline

\end{tabular}
\vspace{-3mm}
\caption{\small Results for entity extraction on conversational corpora. Number of utterances containing each entity type is shown in the first row. F1 scores (multiplied by 100) are displayed. ``-" indicates that result is unavailable. Statistically significant improvements ($p < 0.01$) over all compared systems except Heck et al.~\cite{HeckEtAl2013_interspeech} are in \textbf{bold}. $\dagger$ denotes statistical significance ($p < 0.01$) over Baseline 1$\&$ Baseline 2.}
\label{dialog_entType_tab}
\end{table*}

\begin{table}[ht]
\centering
\footnotesize
\begin{tabular}{|l|c|}
\hline
& \textbf{Recall on Entity Extraction} \\\hline
Baseline 1 (Gazetteer Only) & 84.00\\\hline
Baseline 2 (No Session) & 86.34\\\hline
CRF (mention + type) & 87.33\\\hline
Session model (CRF) & 88.06$\dagger$\\\hline
Session model (CRF with missing label) & \textbf{90.38}\\\hline
\end{tabular}
\vspace{-4mm}
\caption{\small Results for entity extraction on Search log. Recall is reported to measure the portion of entities (including entities that are unseen during training) that are identified. Our first session model (with $\dagger$) achieves statistically significantly ($p<0.01$) better result than the two baselines under a paired-$t$ test. Our model based on CRF trained with missing labels significantly outperforms all the other compared methods.}
\label{web_entity_tab}
\vspace{-3mm}
\end{table}

\subsection{Results: Entity and Relation Extraction}
Due to the large size of the test data, we automatically acquire labels by matching queries and entities with names longer than two words in gazetteer. The resulting labels are taken as the gold standard for evaluation. Because some predictions from our classifier are outside the gazetteer, which makes the precision underestimated, we only report recalls for measuring how many entities or relations are retrieved by the models.
In the test phase, we assume our models only have access to the part of the gazetteer observed in training. 


\noindent \textbf{Comparisons.}
We compare with 1) two baselines, and 2) a linear-CRF that predicts the entity spans and their types simultaneously~\cite{Pound:2012:IKQ:2396761.2396803}. Baseline 1 (Gazetteer Only) is constructed by matching the entities in gazetteer to the queries, where longest matching is selected if one token is mapped to multiple entities. For Baseline 2 (No Session), we follow~\cite{Curran:2003:LIN:1119176.1119200} to construct a multiclass Maximum Entropy classifier with one v.s. all strategy.

\noindent \textbf{Discussion.}
Table~\ref{web_entity_tab} shows that our two session-based models both significantly ($p<0.01$) outperform the two strong baselines under a paired-$t$ test. Our model based on CRF trained with missing labels also produces better performance than all the compared methods\footnote{We do not correct the spelling errors in the queries.}.


For relation extraction, we also compare with the baseline formulated by matching query to relation instances in the gazetteer. Table~\ref{web_relation_tab} presents the results for discovering relation instances for templates \textbf{E-R-E} and \textbf{T-R-E}, where our model outperforms the baseline.

\vspace{-2mm}
\section{Entity Extraction in Spoken Dialog Systems}
\label{dialogManage}
%

\noindent \textbf{SDS Corpora.} To show how our models effectively leverage web search logs for entity extraction in SDS, two corpora are collected internally from real-use scenarios of two SDS, where the users interact with a system by voice and gesture to perform tasks like browsing the internet, searching, querying, and playing. The first conversational browsing dataset is constructed for various domains, and is called \textbf{CB Data} henceforth. The other dataset is designed for a natural language film search application based on Netflix (\textbf{Netflix Data}). As our models are trained on web search log in the \textit{film} domain, we test them on the sessions of film-related tasks with manual transcripts. Specifically, we retain sessions annotated with entities in the film domain.
In total, we have 6691 utterances from 136 sessions from CB; 13851 utterances from 1895 sessions from Netflix. We thus identify the entities and their types on utterance level.


\noindent \textbf{Comparisons.}
We use the gazetteer collected from Section~\ref{webSearch} to construct the same baselines (i.e. Baseline 1 (Gazetteer Only), and Baseline 2 (No Session)). In addition, we compare with the state-of-the-art approach designed for the same task in dialog management from~\cite{HeckEtAl2013_interspeech}, where they elicit training data from Wikipedia articles instead of web queries. They employ a linear-chain CRF to predict entities spans and their types simultaneously.

\noindent \textbf{Discussion.}
Table~\ref{dialog_entType_tab} demonstrates that our session-based models trained on web search logs outperform two nontrivial baselines and the state-of-the-art method in most entity types on spoken dialogs. This shows that our session-based models is able to effectively leverage the behavioral patterns mined from web search log and multi-turn context to improve the entity extraction performance in spoken dialog systems. The performance on genre detection is relatively low. This may due to the different context people use for genres. For example, ``Disney" is a genre in ``Disney movie", but is a place name in ``Disney world". Coreference resolution is not addressed in this work, but we believe it could further improve the performance.


\vspace{-2mm}
\section{Conclusion}
\vspace{-2mm}
In conclusion, we present a full pipeline to leverage semantic web search and browse sessions for a semantic parsing problem in multi-turn spoken dialog systems. Our models are able to apply the learned behaviorial browsing patterns from web seach logs to develop spoken dialog systems. With distant supervision technique, our models achieves better performance than nontrivial baselines and the state-of-the-art methods on both web search queries and spoken dialogs on tasks of entity and relation extraction.


{\small
\bibliographystyle{IEEEbib}

}

\end{document}